\documentclass[letterpaper, 10 pt, conference]{ieeeconf}
\IEEEoverridecommandlockouts                              
\usepackage{times}
\usepackage{epsfig}
\usepackage{graphicx}
\usepackage{amsmath}
\usepackage{amssymb}

\usepackage{graphicx}
\usepackage{epsfig}
\usepackage{epic,eepic}
\usepackage{times}
\usepackage{mathptmx}
\usepackage{amsfonts}
\usepackage{amsmath}
\usepackage{cite}
\usepackage{color}

\usepackage{times}

\definecolor{red}{rgb}{1,0,0}
\definecolor{green}{rgb}{0,1,0}
\definecolor{blue}{rgb}{0,0,1}
\definecolor{violet}{rgb}{1,0,1}
\definecolor{cyan}{cmyk}{1,0,0,0}
\definecolor{magenta}{cmyk}{0,1,0,0}
\definecolor{yellow}{cmyk}{0,0,1,0}

\definecolor{white}{rgb}{1,1,1}

\usepackage{jumoline}

\newcommand{\CO}[1]{}

\newcommand{\CommentOut}[1]{}

\newcommand{\noeditage}[1]{#1} \newcommand{\editage}[1]{}

\begin{document}

\newcommand{\FIGcenter}[4]{
  \begin{minipage}{#1\linewidth}
    \begin{center}
      \includegraphics[bb=0 0 #3 #4, clip, width=\linewidth]{#2}\\
    \end{center}
  \end{minipage}
}
\newcommand{\FIG}[3]{
\begin{minipage}[b]{#1cm}
\begin{center}
\includegraphics[width=#1cm]{#2}\\
{\scriptsize #3}
\end{center}
\end{minipage}
}

\newcommand{\FIGU}[3]{
\begin{minipage}[b]{#1cm}
\begin{center}
\includegraphics[width=#1cm,angle=180]{#2}\\
{\scriptsize #3}
\end{center}
\end{minipage}
}

\newcommand{\FIGm}[3]{
\begin{minipage}[b]{#1cm}
\begin{center}
\includegraphics[width=#1cm]{#2}\\
{\scriptsize #3}
\end{center}
\end{minipage}
}

\newcommand{\FIGR}[3]{
\begin{minipage}[b]{#1cm}
\begin{center}
\includegraphics[angle=-90,clip,width=#1cm]{#2}
\\
{\scriptsize #3}
\vspace*{1mm}
\end{center}
\end{minipage}
}

\newcommand{\FIGRpng}[5]{
\begin{minipage}[b]{#1cm}
\begin{center}
\includegraphics[bb=0 0 #4 #5, angle=-90,clip,width=#1cm]{#2}\vspace*{1mm}
\\
{\scriptsize #3}
\vspace*{1mm}
\end{center}
\end{minipage}
}

\newcommand{\FIGpng}[5]{
\begin{minipage}[b]{#1cm}
\begin{center}
\includegraphics[bb=0 0 #4 #5, clip, width=#1cm]{#2}\vspace*{-1mm}\\
{\scriptsize #3}
\vspace*{1mm}
\end{center}
\end{minipage}
}

\newcommand{\FIGtpng}[5]{
\begin{minipage}[t]{#1cm}
\begin{center}
\includegraphics[bb=0 0 #4 #5, clip,width=#1cm]{#2}\vspace*{1mm}
\\
{\scriptsize #3}
\vspace*{1mm}
\end{center}
\end{minipage}
}

\newcommand{\FIGRt}[3]{
\begin{minipage}[t]{#1cm}
\begin{center}
\includegraphics[angle=-90,clip,width=#1cm]{#2}\vspace*{1mm}
\\
{\scriptsize #3}
\vspace*{1mm}
\end{center}
\end{minipage}
}

\newcommand{\FIGRm}[3]{
\begin{minipage}[b]{#1cm}
\begin{center}
\includegraphics[angle=-90,clip,width=#1cm]{#2}\vspace*{0mm}
\\
{\scriptsize #3}
\vspace*{1mm}
\end{center}
\end{minipage}
}

\newcommand{\FIGC}[5]{
\begin{minipage}[b]{#1cm}
\begin{center}
\includegraphics[width=#2cm,height=#3cm]{#4}~$\Longrightarrow$\vspace*{0mm}
\\
{\scriptsize #5}
\vspace*{8mm}
\end{center}
\end{minipage}
}

\newcommand{\FIGf}[3]{
\begin{minipage}[b]{#1cm}
\begin{center}
\fbox{\includegraphics[width=#1cm]{#2}}\vspace*{0.5mm}\\
{\scriptsize #3}
\end{center}
\end{minipage}
}

\newcommand{\figGRID}{
\begin{figure}[h]
\begin{center}
\FIGpng{8}{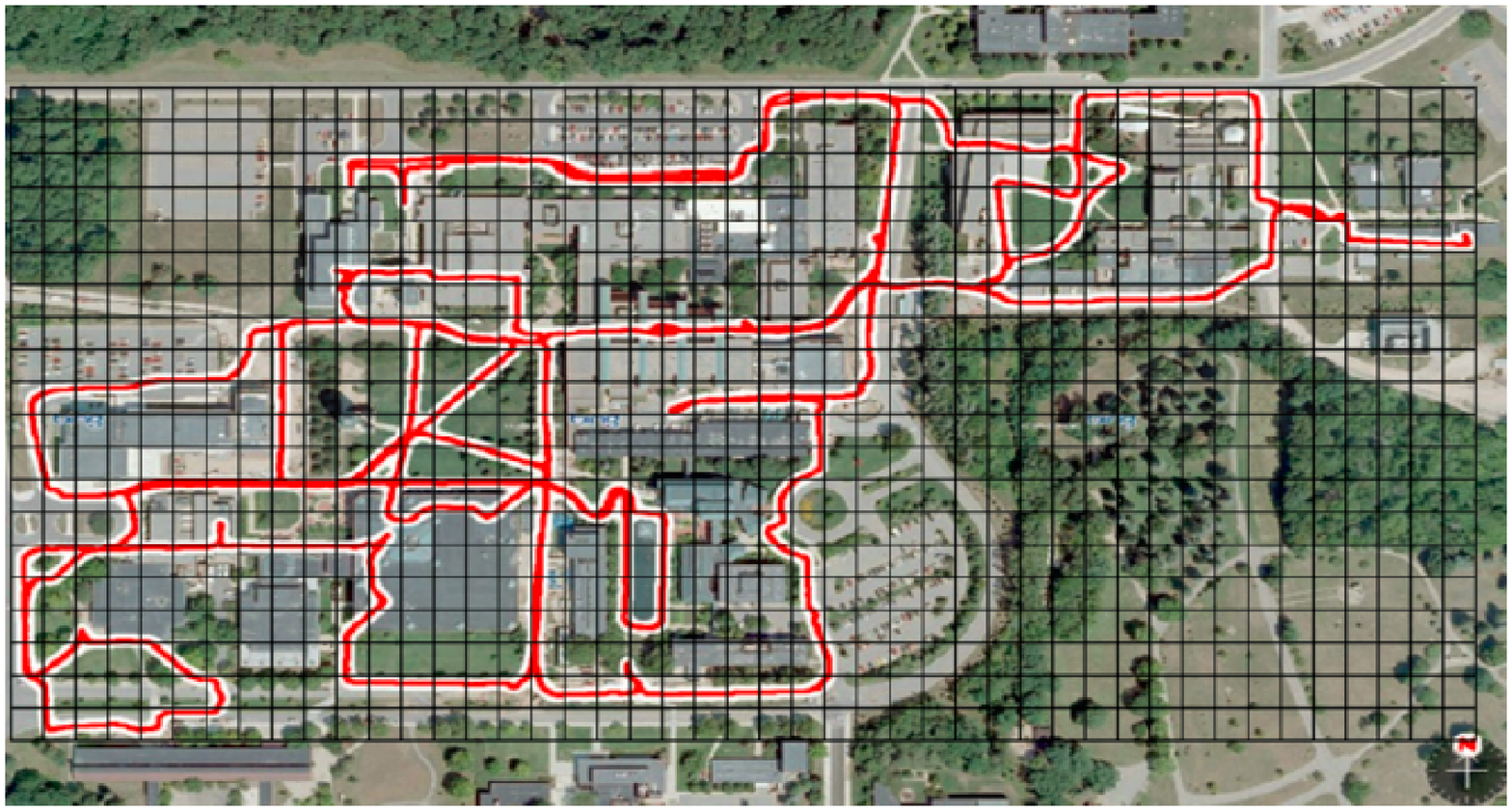}{}{668}{355}
\caption{NCLT dataset and definition of place classes. Each place class corresponds to 20$\times$20 $[m^2]$ grid cell.}\label{fig:env}
\end{center}
\end{figure}
}

\newcommand{\vs}{\hspace*{-1mm}}

\newcommand{\SW}[2]{#1}

\renewcommand{\CO}[2]{#2}

\onecolumn

\title{\LARGE \bf
Long-Term Vehicle Localization
by Recursive Knowledge Distillation
}
\author{Hiroki Tomoe ~~~~~~~ Tanaka Kanji
\thanks{Our work has been supported in part by 
JSPS KAKENHI 
Grant-in-Aid 
for Scientific Research (C) 26330297, and (C) 17K00361.}
\thanks{The authors are with University of Fukui, Japan. 
{\tt\small \{hiroki, tnkknj\}@u-fukui.ac.jp}}
}
\maketitle

\newcommand{\codeA}{
\begin{figure}[t]
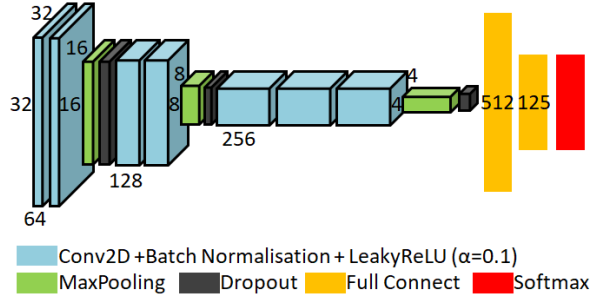

\fbox{
\begin{minipage}[b]{8cm}
{\scriptsize
x = Convolution2D(64, (3, 3), padding='same')(input\_layer) \\
x = BatchNormalization()(x) \\
x = advanced\_activations.LeakyReLU(alpha=0.1)(x) \\
x = Convolution2D(64, (3, 3), padding='same')(x) \\
x = BatchNormalization()(x) \\
x = advanced\_activations.LeakyReLU(alpha=0.1)(x) \\
x = MaxPooling2D((2, 2), strides=(2, 2))(x) \\
x = Dropout(0.3)(x) \\
x = BatchNormalization()(x) \\
x = advanced\_activations.LeakyReLU(alpha=0.1)(x) \\
x = Convolution2D(128, (3, 3), padding='same')(x) \\
x = BatchNormalization()(x) \\
x = advanced\_activations.LeakyReLU(alpha=0.1)(x) \\
x = MaxPooling2D((2, 2), strides=(2, 2))(x) \\
x = Dropout(0.3)(x) \\
x = Convolution2D(256, (3, 3), padding='same')(x) \\
x = BatchNormalization()(x) \\
x = advanced\_activations.LeakyReLU(alpha=0.1)(x) \\
x = Convolution2D(256, (3, 3), padding='same')(x) \\
x = BatchNormalization()(x) \\
x = advanced\_activations.LeakyReLU(alpha=0.1)(x) \\
x = Convolution2D(256, (3, 3), padding='same')(x) \\
x = BatchNormalization()(x) \\
x = advanced\_activations.LeakyReLU(alpha=0.1)(x) \\
x = MaxPooling2D((2, 2), strides=(2, 2))(x) \\
x = Dropout(0.3)(x) \\
x = Flatten()(x) \\
x = Dense(512, activation=None)(x) \\
x = BatchNormalization()(x) \\
x = advanced\_activations.LeakyReLU(alpha=0.1)(x) \\
logits = Dense(num\_classes, activation=None)(x) \\
output = Activation('softmax')(logits) \\
}
\end{minipage}
}
\caption{CNN architecture.}\label{fig:architecture}
\end{figure}
}

\newcommand{\codeB}{
\begin{figure}[t]
\fbox{
\begin{minipage}[b]{8cm}
{\scriptsize
opt = keras.optimizers.Adam(lr=0.003, beta\_1=0.9, beta\_2=0.999, epsilon=1e-08) \# Adam Optimizer \\
}
\end{minipage}
}
\caption{Transfer from the current season's visual experience.}\label{fig:learningfromdata}
\end{figure}
}

\newcommand{\codeC}{
\begin{figure}[t]
\fbox{
\begin{minipage}[b]{8cm}
{\scriptsize
logits\_T = Lambda(lambda x: x/10.0, name="logits")(logits) \\
output = Activation("softmax")(logits\_T) \\
opt = keras.optimizers.Adam(lr=0.003, beta\_1=0.9, beta\_2=0.999, epsilon=1e-08)
\# Adam Optimizer \\
}
\end{minipage}
}
\caption{Transfer from previous seasons' teacher CNNs.}\label{fig:transferfromcnn}
\end{figure}
}

\newcommand{\tabST}{
\begin{table}[t]
\begin{center}
\caption{TSA Strategies.}\label{tab:st}
\begin{tabular}{|lc|lc|}
\hline
\#1 & $A_4([1,1],1.0)$ &
\#2 & $A_4([2,2],0.5)$ \\ \hline
\#3 & $A_4([2,2],1.0)$ & 
\#4 & $A_4([3,3],1.0)$ \\ \hline
\#5 & $A_4([1,2],1.0)$ & 
\#6 & $A_4([0,2],1.0)$ \\ \hline
\#7 & $B_4$ & 
\#8 & $A_4([0,4],1.0)$ \\ \hline
\#9 & $B_2 + A_2([0,0],1.0)$ &
\#10 & $B_2 + A_2([2,2],1.0)$ \\ \hline
\#11 & 
\multicolumn{3}{c|}{
$B_2 + A_1([1,1],1.0) + A_1([1,1],1.0)$ 
} \\ \hline
\#12 & $B_2 + A_2([1,1],1.0)$ &
\#13 & $C_{4,2}$ \\ \hline
\#14 & \multicolumn{3}{c|}{
$A_4([0,0],1.0)$ 
} \\ \hline
\end{tabular}
\end{center}
\end{table}
}

\newcommand{\figFIRST}{
\begin{figure}
\begin{center}
\FIG{8}{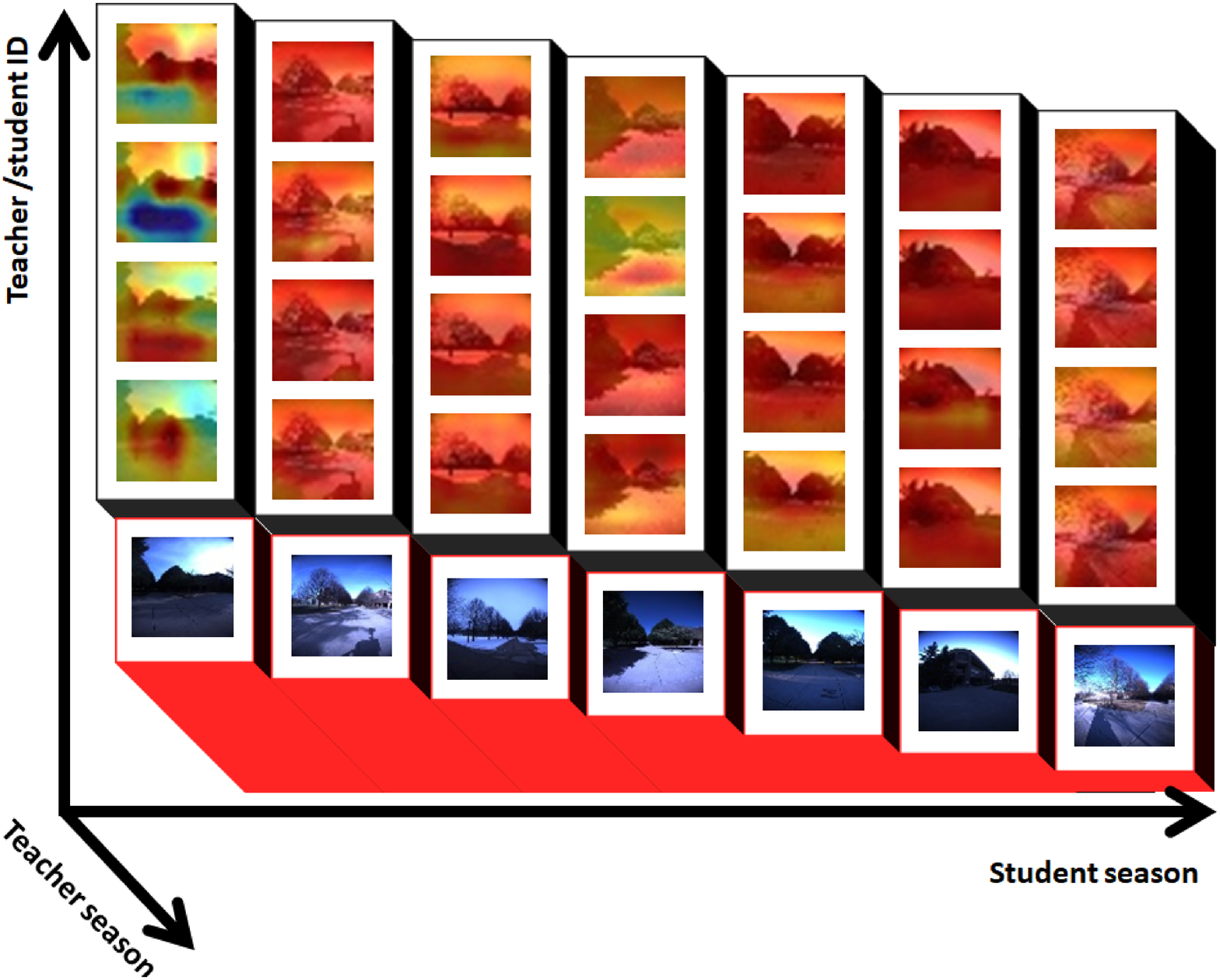}{}
\caption{
We 
address the 
task of 
sequential multi-domain 
visual place recognition (SMD-VPR) \cite{FangITSC18}
from the novel perspective of 
multi-teacher multi-student 
knowledge distillation
(MTMS-KD) \cite{MTMSKD}.
For each $t$-th season ($t$-th column),
the teacher set consists
of $K$ previous seasons CNNs (1st, 2nd, 3rd and 4th rows),
and a tentative CNN that is pretrained on the current season's training set (5th row).
We 
present a novel 
recursive knowledge distillation (RKD)
framework 
that recursively compresses all the previous seasons' knowledge (i.e., 1st, $\cdots$, $(t-1)$-th columns)
into a constant size CNN ensemble (i.e., $t$-th column).
}\label{fig:figFIRST}
\end{center}
\end{figure}
}

\newcommand{\figAT}{
\begin{figure}
\begin{flushleft}
\hspace*{3cm}\FIG{12}{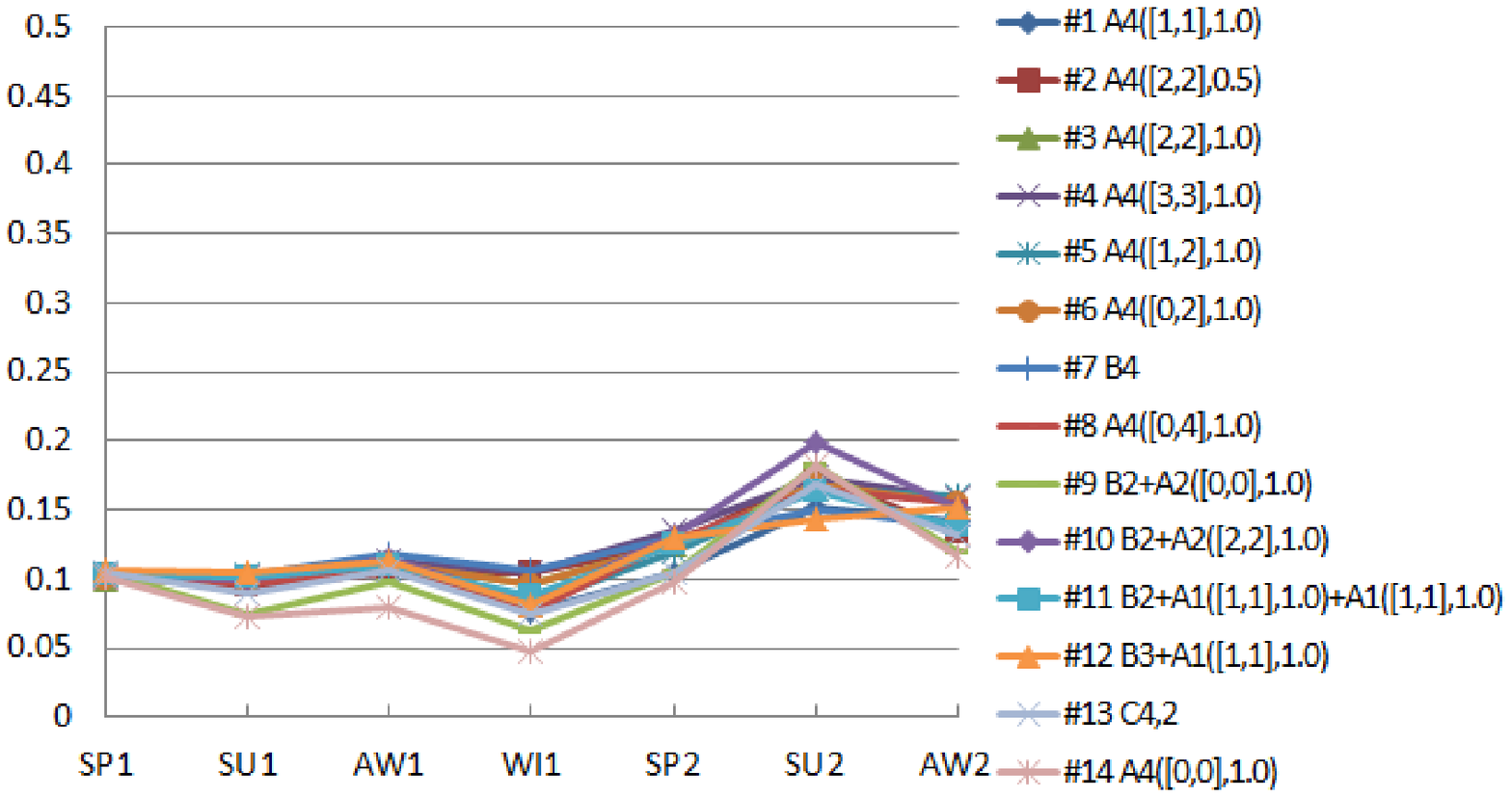}{}
\caption{Top-1 accuracy.}\label{fig:at}
\end{flushleft}
\end{figure}
}

\newcommand{\figF}{
\begin{figure}
\begin{center}
\FIG{8}{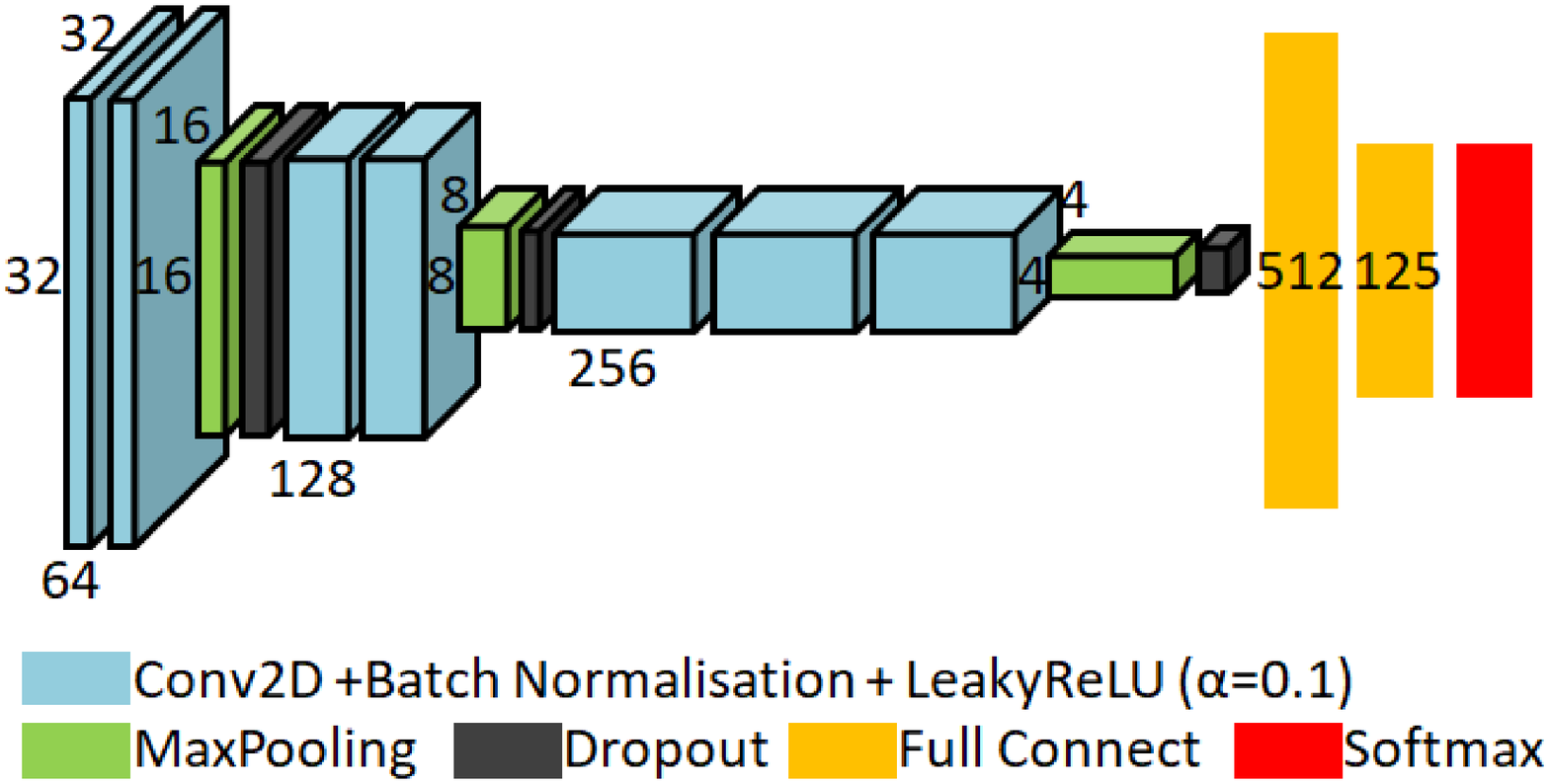}{}
\caption{CNN architecture.}\label{fig:F}
\end{center}
\end{figure}
}

\section*{\centering Abstract}
\textit{
Most of the current state-of-the-art frameworks for cross-season visual place recognition (CS-VPR) focus on domain adaptation (DA) to a {\it single} specific season. From the viewpoint of long-term CS-VPR, such frameworks do not scale well to {\it sequential multiple domains} (e.g., spring $\rightarrow$ summer $\rightarrow$ autumn $\rightarrow$ winter $\rightarrow$ $\cdots$). The goal of this study is to develop a novel long-term ensemble learning (LEL) framework that allows for a constant cost retraining in long-term sequential-multi-domain CS-VPR (SMD-VPR), which only requires the memorization of a small constant number of deep convolutional neural networks (CNNs) and can retrain the CNN ensemble of every season at a small constant time/space cost. We frame our task as the multi-teacher multi-student knowledge distillation (MTMS-KD), which recursively compresses all the previous season's knowledge into a current CNN ensemble. We further address the issue of teacher-student-assignment (TSA) to achieve a good generalization/specialization tradeoff. Experimental results on SMD-VPR tasks validate the efficacy of the proposed approach.
}

\section{Introduction}

Visual recognition
of places across different seasons  has been a central challenge 
in long-term autonomy
called cross-season visual place recognition (CS-VPR).
One of major sources 
of difficulty is 
the change in appearance of place 
caused by domain shifts, such as cyclic seasonal changes, day-night illumination changes, and structural changes \cite{SeqSlamOrg}.

Most of the 
current state-of-the-art CS-VPR
frameworks focus on domain adaptation (DA) to a {\it single} specific season.
One of the predominant approaches is 
deep learning (DL) -based convolutional neural networks (CNN) \cite{alexnet},
which adopts a CNN as a visual place classifier (VPC)
to a specific season's training images.

From the viewpoint of long-term CS-VPR, such methods do not scale well to {\it sequential multiple domains} (e.g., spring $\rightarrow$ summer $\rightarrow$ autumn $\rightarrow$ winter $\rightarrow$ $\cdots$).
They require a vehicle to explicitly store a number of CNNs for a long period of time over sequential multiple seasons and a number of training images proportional to the number of experienced seasons/places. This severely limits the scalability of the algorithm in both time and memory space.

Our goal is to allow for a constant cost retraining
in long-term CS-VPR
across sequential multi-domain.
We only require to memorize a small constant number of CNNs.
We can retrain the CNN ensemble 
for every season at a small constant time/space cost.

We frame our task as a knowledge distillation (KD) problem with constraints of appearance similarity over different seasons/places.
More specifically, we model the previous and current season's CNNs as teachers and students, respectively,
based on the recently developed 
multi-teacher multi-student KD (MTMS-KD) \cite{MTMSKD}.
A key advantage
of the algorithm is that 
we only require
a constant number of previous season's teachers
and current seasons's training data
as prior knowledge
and the space/time similarities between places/seasons
as the constraints between different domains.

We propose a new recursive KD (RKD) algorithm, 
whereby the current $t$-th season's training images $D^t$ and the previous $(t-1)$-th season's CNN ensemble $S^{t-1}$ are compressed into a new CNN ensemble $S^t$ (Fig. \ref{fig:figFIRST}).
It is worth noting
that
this is a recursive 
compression procedure
that 
the compression of
all the past seasons' training images
$D^1$, $\cdots$, $D^t$
into the 
current 
constant-size 
CNN ensemble $S^t$.

\noeditage{
\figFIRST
}

We further address a key question in RKD, termed the teacher-student assignment (TSA) problem: ``which student should be trained by which teacher?" The TSA problem itself is difficult and it involves a generalization/specialization tradeoff: if a student's CNN is trained by a specific season's teacher, its season-specific VPR ability will increase, while its generic VPR ability will decrease. Thus, we have two possible choices: either we train a specific student's CNN with a specific teacher's CNN or we do not, and we have exponential number of possible choices to the number of experienced seasons. Searching the best choice in this large number of possible TSAs is often unfeasible. We explore and discuss several strategies for the TSA problem.

The main contributions of our work are as follows:
\begin{enumerate}
\item
We propose a novel sequential-multi-domain CS-VPR (SMD-VPR)
framework,
which
requires 
a constant cost for long-term memory and retraining.
\item
We frame the SMD-VPR task as 
a recursive KD (RKD)
and address the novel 
TSA problem.
\end{enumerate}

In this study, we formulate 
the SMD-VPR
as a CNN-based classification problem,
and focus on the issue of long-term RKD and TSA. Further, in this study we extend our previously developed long-term ensemble learning (LEL) framework
\cite{FeiJACIII18,FangITSC18}
to a memory efficient MTMS-KD system.
In \cite{FeiJACIII18},
we developed
a recursive
MTMS framework
whereby teachers' CNNs in the previous season
are directly retrained into the current season's CNNs,
which can then be used as student CNNs.
In \cite{FangITSC18},
we addressed
the issue
of space efficiency in the MTMS framework  
for the first time,
and
presented
an efficient MTMS framework
by representing a CNN
as a collection of 
season-specific training images.
However,
such frameworks
require to explicitly store
a number of training images
proportional to the number of experienced seasons and places,
which severely limits the scalability
to long-term large-size SMD-VPR tasks.

\section{Approach}

The long-term map learning framework consists of two alternately repeated missions (one iteration per season): exploration and adaptation. 
The system is initialized with a size one classifier set $S^1$$=\{c_1^1\}$,
which consists of a single CNN
classifier $c^1_1$ that is obtained 
by pretraining a CNN using the first season's training data. 
A new classifier set
$S^t=$$\{c_j^t\}_{j=1}^{K_t}$ is then obtained 
by a teacher-to-student KD in each $t$-th iteration ($t> 1$). 
In experiments, for the initial CNN classifier $c_1^1$,
we use the CNN architecture pretrained on the CIFAR10 dataset, and we consider one iteration of the two missions per season.

The exploration mission aims at 
maximum possible vehicle exploration of the entire environment, while keeping track of the vehicle's global position (e.g., using pose tracking \cite{DeepVO18} and relocation \cite{ibowlcd}), 
to collect mapped images that have global viewpoint information, and optionally, the collected data may be further post-processed to refine the viewpoint information by structure-from-motion (SfM) \cite{StructureFromMotionNew} or SLAM \cite{engel2014lsd}. 

All the collected images that have viewpoint information can be used as training data for the subsequent $t$-th adaptation mission. 
We denote training data that is collected in the $t$-th exploration as $D^t=\{(v, I)\}$, where $I$ and $v$ are an image and its viewpoint, respectively.

Given a set of previous season's $K^{t-1}$ teacher CNNs
and
the current season's train set,
the objective of MTMS-KD is 
to train a new set of current season's $K^{t}$ student CNNs.

Although our approach is sufficiently general and applicable to generic variable size teacher/student CNN sets,
in this study,
we focus on simple
constant size CNN sets (i.e., $K^t=K^{t-1}$)
with $K=4$.

To solve this problem,
we develop two different types of algorithms:
(1)
the TSA algorithm 
which 
assigns best teachers
to each student, and
(2)
the RKD algorithm
which
trains a new set of current season's $K$ student CNNs,
from
the previous season's $K$ teacher CNNs
and the current season's training images,
given a specific teacher-student-assignment.

\subsection{Performance Metric}

The performance metric of a VPR system
is based on the
top-$X$ accuracy.
In it,
the CNN ensemble is modeled as a ranking function,
which outputs
a ranked list of
place classes
in descending order of
relevance score (e.g., confidence).
Then,
the accuracy of top-$X$
in the ranked list
is evaluated
with respect to the ground-truth viewpoint
obtained by GPS measurements.

Because the both subtasks are highly ill-posed and no ground-truth solutions (GTs) are available,
we
investigate 
the relative performance of
metrics
that do not rely on ground-truth TSA and RKD,
but only require GT viewpoint information.
To evaluate the long-term SMD-VPR performance,
the above top-$X$ accuracy
is evaluated
by using the latest up-to-date CNN ensemble.
Naturally,
the evaluation is performed
after 
rather than before
the $t$-th adaptation mission 
in each $t$-th season.

To evaluate the generalization ability of our VPR system,
we use the next season's unseen data
as the test data for the current season's ensemble.
Note that
this means that
students are often tested with
test data of
unseen seasons,
which the students and previous teachers 
have never experienced before.
This makes our CS-VPR
a very challenging problem.

\noeditage{
\tabST

\figF

}

\subsection{TSA}\label{sec:tsa}

The TSA task aims to achieve a good balance between knowledge transferring from current season's teacher and previous seasons' teachers.
However,
finding a way to achieve such a balance
is not straightforward.
Instead,
we have developed several possible strategies
and investigated their effectiveness.
In this study,
we consider 14 different strategies \#1, $\cdots$, \#14, 
as listed in 
Table \ref{tab:st}.
In these strategies,
$A_s$ and $B_s$
are two different functions
that 
take a size $s$ $(s\in [1,K])$ collection of the previous season's CNNs $S^{t-1}=$$\{c_1^{t-1}$, $\cdots$, $c_s^{t-1}\}$, 
and outputs a 
$(s+1)\times s$
binary matrix
called the TSA matrix $M=\{M_{ij}\}$.
An assignment 
$M_{ij}$ 
is represented by a connection between
a $i$-th teacher CNN 
and 
for the $j$-th student CNN.
A teacher CNN can be either 
a previous season's CNN $c^{t-1}_i$$(i>0)$ (denoted as ``previous teacher")
or a tentative CNN $c^t_0$ that is pretrained on the current season's training set (denoted as ``current teacher").
We use an identical CNN architecture 
for all the CNNs
for previous and current season's teachers.
A TSA $M_{ij}$
for $j$-th student CNN
represents
the assignment of the previous teacher $c^{t-1}_{i}$,
where
$M_{0j}=1$
represents 
the assignment of the current teacher $c^t_0$,
and
$M_{ij}=1 (i>0)$
represents
the assignment of the previous $i$-th teacher $c^{t-1}_{a_i}$.
The main difference between the set functions $A_s$ and $B_s$
lies in the way the number of teachers are determined.
The number of previous season's teachers is dependent on the number of current season's teachers for the function $A_s$, while it is independent for the functions $B_s$.
The function $A_s([a,b],p)$ samples the number of previous season's teachers in the range $[a,b]$, and determines whether to use the current season's teacher with the probability of $p$.
The function $B_s$ does not distinguish the two teacher types (i.e., previous or current) but simply samples the total number of teachers in range $[1,s]$.
The function $C_{s,s'}$ $(s'<s)$ splits the students into the groups $\{1,\cdots,s'\}$, and $\{s'+1,\cdots,s\}$, for the former group, it assigns previous and/or current teachers with the number of teachers in range $[1,5]$, while for the latter group, it only assigns the current season's teacher (i.e., $i=0$).
It is worth noting that the method \#14
enforces the use of only current season's teachers.
It is viewed as the baseline method in the experimental section \ref{sec:exp}.

\subsection{Training Procedure}

We use an identical CNN architecture for all the teachers and student CNNs. The CNN architecture is based on the one in the CIFAR10 \cite{krizhevsky2010convolutional}. It consists of 16 layers and the LeakyReLU activation function is employed for each layer. 
Fig. \ref{fig:F} illustrates the CNN architecture.
For the distillation, the training data consists of 25,966 samples from a dataset ``2012/5/26" which is independent of training/test data used in experiments. The Adam optimizer \cite{kingma2014adam} is used with a learning rate of 0.003, $\beta_1=0.9$, $\beta_2=0.999$, $\epsilon=10^{-8}$, and a weight decay of 0. For the loss function, the following soft loss is used.
\begin{equation}
Loss_{softmax} = - \sum_i p_i \log (q_i),
\end{equation}
where
\begin{equation}
q_i = \frac{ exp(z_i/T) }{ \sum_j exp(z_j/T) },
\end{equation}
\begin{equation}
p_i = \frac{ exp(v_i/T) }{ \sum_j exp(v_j/T) },
\end{equation}
where 
$z_i$ and $v_i$ are 
the $i$-th class logits from the student and teacher CNNs, respectively.
The temperature $T$ is set 
to 1 for pretraining
and
to 10 for KD.

\noeditage{
\figGRID
\figAT
}

\section{Experiments and Discussions}\label{sec:exp}

We evaluated the proposed SMD-VPR framework on a challenging cross-season dataset: a public NCLT dataset \cite{nclt}.
The NCLT 
is a large-scale, long-term autonomy dataset for robotics
research that was collected at the University of Michigan's
North Campus by a Segway vehicle platform. The data we
used in this study includes view image sequences along the
vehicle trajectories acquired by the front facing camera of the
Ladybug3 platform (Fig. \ref{fig:env}). 
Specifically,
we used a length 8 sequence of datasets:
``2012/3/31 (SP1)",  ``2012/8/4 (SU1)",  ``2012/11/17 (AU1)", ``2012/1/22 (WI1)",  ``2012/5/11 (SP2)", ``2012/8/20 (SU2)", ``2012/11/16 (AU2)", and ``2012/1/8 (WI2)".
For each dataset,
we impose a regular 20$\times$20 $[m^2]$ grid,
and view each cell as a place class candidate.
The number of training/testing images are very different between different place cells and is dependent on the viewpoint trajectories followed by the vehicle for each dataset.
We ignore the place cells that have an insufficient number of training/testing images for at least one dataset.
Consequently, 125 place cells are determined to be valid 
and are used for our VPR tasks.

Figure \ref{fig:at}
shows the performance of each method in each season.
The performance in terms of 
Top-$X$($X$=1, 5, 10)
accuracy is investigated for
two standard 
ensemble strategies,
namely
``averaging" and ``merge+sort".
The method \#14 is the baseline method as described in Section \ref{sec:tsa}.
One can see
that the proposed methods outperformed the baseline method 
in most of the experiments.

\bibliographystyle{IEEEtran}
\bibliography{cite}

\end{document}